\documentclass[conference]{IEEEtran}
\newcommand\tab[1][1cm]{\hspace*{#1}}
\IEEEoverridecommandlockouts
\usepackage{cite}
\usepackage{amsmath,amssymb,amsfonts}
\usepackage{algorithmic}
\usepackage{graphicx}
\usepackage{textcomp}
\usepackage{adjustbox}
\usepackage{xcolor}
\usepackage{multirow}
\def\BibTeX{{\rm B\kern-.05em{\sc i\kern-.025em b}\kern-.08em
    T\kern-.1667em\lower.7ex\hbox{E}\kern-.125emX}}
\DeclareMathOperator*{\argmax}{\arg\max}

\begin{document}

\title{Prediction of Progression to Alzheimer's disease with Deep InfoMax
\thanks{Corresponding author: Alex Fedorov, afedorov@mrn.org.
**Data used in the preparation of this article were obtained from the Alzheimer's Disease Neuroimaging Initiative (ADNI) database (adni.loni.usc.edu). As such, the investigators within the ADNI contributed to the design and implementation of ADNI and/or provided data but did not participate in analysis or writing of this report. }
}

\author{\IEEEauthorblockN{Alex Fedorov\IEEEauthorrefmark{1}\IEEEauthorrefmark{2}, R Devon Hjelm\IEEEauthorrefmark{3}\IEEEauthorrefmark{4}\IEEEauthorrefmark{5},
Anees Abrol\IEEEauthorrefmark{1}\IEEEauthorrefmark{2}, Zening Fu\IEEEauthorrefmark{1}, Yuhui Du\IEEEauthorrefmark{1}\IEEEauthorrefmark{6}, Sergey Plis\IEEEauthorrefmark{1}, Vince D. Calhoun\IEEEauthorrefmark{1}\IEEEauthorrefmark{2}\\and for the Alzheimer's Disease Neuroimaging Initiative**}
\IEEEauthorblockA{\IEEEauthorrefmark{1}The Mind Research Network, Albuquerque, New Mexico, USA }
\IEEEauthorblockA{\IEEEauthorrefmark{2}Department of Electrical and Computer Engineering, University of New Mexico, Albuquerque,
New Mexico, USA}
\IEEEauthorblockA{\IEEEauthorrefmark{3}Microsoft Research, Montreal, Canada}
\IEEEauthorblockA{\IEEEauthorrefmark{4}Montreal Institute for Learning Algorithms, Montreal, Canada}
\IEEEauthorblockA{\IEEEauthorrefmark{5}Department of Computer Science and Operations Research, University of Montreal, Montreal, Canada}
\IEEEauthorblockA{\IEEEauthorrefmark{6}School of Computer \& Information Technology, Shanxi University, Taiyuan, China}
}

\maketitle

\begin{abstract}
Arguably, unsupervised learning plays a crucial role in the majority of algorithms for processing brain imaging. A recently introduced unsupervised approach Deep InfoMax (DIM) is a promising tool for exploring brain structure in a flexible non-linear way. In this paper, we investigate the use of variants of DIM in a setting of progression to Alzheimer's disease in comparison with supervised AlexNet and ResNet inspired convolutional neural networks. As a benchmark, we use a classification task between four groups: patients with stable, and progressive mild cognitive impairment (MCI), with Alzheimer's disease, and healthy controls. Our dataset is comprised of 828 subjects from the Alzheimer’s Disease Neuroimaging Initiative (ADNI) database. Our experiments highlight encouraging evidence of the high potential utility of DIM in future neuroimaging studies.
\end{abstract}

\begin{IEEEkeywords}
CNN, MRI, Deep InfoMax, classification, unsupervised
\end{IEEEkeywords}

\section{Introduction}
According to \cite{trautmann2016economic}, the economic costs of mental disorders have the highest impact on economic growth, direct and indirect costs and the statistical value of life. 
One essential tool for better understanding mental illness is to use noninvasive neuroimaging (e.g., structural magnetic resonance imaging (MRI) images) along with machine learning to learn brain structure.

Deep Learning has been integral to the successes of machine learning for numerous demanding real-world applications, e.g., state-of-the-art image classification~\cite{huang2017densely} and self-driving cars~\cite{xiong2019upsnet}. 
While many of Deep Learning's successes involve supervised learning, supervised approaches can fail when data annotation (e.g., labels) is limited or unavailable.
When there is sufficient data, supervised models can not only perform well on holdout sets but provide representations that generalize well to other supervised settings~\cite{zhang2017learning}.
However, when there is insufficient data, a supervised learner tends to discriminate on low-level (e.g., pixel-level, \emph{trivial}) information, which hurts generalization performance.
A model that generalizes well needs to extract meaningful high-level information (e.g., a collection of important features at the input level).
In order to address this, many successful applications of machine learning to neuroscience rely on unsupervised learning~\cite{calhoun2001method, hjelm2014restricted, castro2016deep, plis2014deep} to extract representations of brain imaging data.
These representations are then used as input to an off-the-shelf classifier (i.e., semi-supervised learning).

However, prior work on unsupervised learning of brain imaging data is either linear or weakly nonlinear~\cite{calhoun2001method, hjelm2014restricted} or are highly restrictive in parameterization~\cite{castro2016deep}, and do not represent flexible methodology for learning representations.

In this work, we explore using DIM~\cite{hjelm2018learning} to learn deep non-linear representations of neuroimaging data as an output of a convolutional neural network.
DIM works by maximizing the mutual information between a high-level feature vector and low-level feature maps of a highly flexible convolutional \emph{encoder} network by training a second neural network that maximizes a lower bound on a divergence (probabilistic measure of difference) between the joint or the product of marginals of the encoder input and output.
The estimates provided by this second network can be used to maximize the mutual information of the features in the encoder with the input. Unlike other popular unsupervised auto-encoding approaches such as VAE~\cite{kingma2013auto}, DIM doesn't require a decoder. Hence it significantly reduces memory requirements of the model for volumetric data.

We evaluate DIM by performing a downstream classification task between four groups: patients with stable and progressive MCI, with Alzheimer's disease and healthy controls, using only the resulting representation from DIM as input to the classifier. We compare DIM to two convolutional networks with AlexNet~\cite{NIPS2012_4824} and ResNet~\cite{he2016deep} inspired architectures trained with supervised learning. On strict evaluation, we show comparable performance to supervised methods and to previously reported~\cite{abrol2018deep, vieira2017using, 6963480, suk2016deep} classification performance.

\section{Materials and Methods}

\subsection{Deep InfoMax}
Let $\mathbf{X} :=\{x^{(i)} \in \mathcal{X}\}$ and $\mathbf{Z}:= \{z^{(i)} \in \mathcal{Z}\} $ be the input and output variables of a neural network encoder, $E_{\phi}: \mathcal{X} \rightarrow \mathcal{Z} $ with parameters $\phi$, where $\mathcal{X}$, and $\mathcal{Z}$ are its domain and range.
We wish to find the parameters that maximize the following objective:
\begin{equation}
    (\hat{\phi}, \hat{\theta})_G 
    = \argmax_{\phi, \theta} \hat{\mathcal{I}}_{\theta}(X;Z),
    \label{eq:infomax}
\end{equation}
where $\hat{\mathcal{I}}_{\theta}$ is the mutual information estimate provided by a different network with parameters $\theta$, and $Z=E_{\phi}(X)$ is the output of the encoder.

A parametric estimator for the mutual information can be found by training a \emph{statistics network} to maximize a lower bound based on the Fenchel-dual~\cite{nowozin2016f} or the Donsker-Varadhan representation~\cite{donsker1983asymptotic, belghazi2018mine} of the Kullback–Leibler divergence $D_{KL}$.
The Donsker-Varadhan-based estimator is a consistent, asymptotically unbiased estimator has been shown to outperform nonparametric estimators, and can also be used to improve deep generative models~\cite{belghazi2018mine}.
However, $D_{KL}$ is unbounded, which can be problematic if the above estimators are used for training deterministic neural network encoders.
\cite{hjelm2018learning} showed that using an estimator based on the Jensen-Shannon divergence (JSD) (i.e., simple binary cross-entropy) is more stable and works well in practice, and it has been shown that this estimator also yields a good estimator for mutual information~\cite{hjelm2018learning, poolevariational}:
\begin{align}
   \label{eq:diim_jsd}
   \begin{split}
   \hat{\mathcal{I}}_{\phi, \theta}^\text{(JSD)}(X; E_{\phi}(X)) :=
   &\mathbb{E}_{\mathbb{P}_X} [-\text{sp}(-T_{\theta}(X, E_{\phi}(X)))] - \\
   &\mathbb{E}_{\mathbb{P}_X \otimes \mathbb{P}_X}[\text{sp}(T_{\theta}(X', E_{\phi}(X)))],
\end{split}
\end{align}
where $T_{\theta}$ is a statistics network with parameters $\theta$, $\text{sp} = \log(1+e^z)$ (softplus function) and $X'$ is another input sampled from the data distribution independently from $X$.
In addition, the Noise-Contrastive variant of the estimator (NCE)~\cite{oord2018representation} was shown to work well in practice~\cite{hjelm2018learning}:
\begin{align}
    \begin{split}
        &\hat{\mathcal{I}}_{\phi, \theta}^\text{(NCE)}
        (X; E_{\phi}(X)) := \\
        &\mathbb{E}_{\mathbb{P}} 
        \left[ 
            T_{\theta}(X, E_{\phi}(X)) -
            \log \sum_{X' \in \mathcal{X}_b} e^{T_{\theta}(X',  E_{\phi}(X))} 
        \right].
    \end{split}
    \label{eq:diim_nce}
\end{align}
Here, $\mathcal{X}_b = \{X\} \bigcup \mathcal{X}_n$ are a set of samples where $\mathcal{X}_n$ are a set of \emph{negative samples} drawn from the data distribution, such that there is exactly one positive example in $\mathcal{X}_b$ ($X$ occurs exactly once).

\cite{hjelm2018learning} showed that maximizing the mutual information between the \emph{complete} input and output of an encoder are insufficient for learning good representations for downstream classification tasks, as this approach can still focus on lower-level ``trivial" localized details.
Instead, they show that maximizing the mutual information between the high-level representation, $Z = E_{\phi}(X)$ and \emph{patches} of an input image can achieve highly competitive results.
The intuition is that this approach encourages the high-level representation to learn information that is \emph{shared} across the input.
It is suitable for many classification tasks, as we expect that class-discriminative features should be evident across many spatial locations of the input.
For a convolutional encoder $E$, the \emph{local} DIM objective can be written in a compact form:
\begin{equation}
    \label{eq:diim-l}
    (\hat{\phi}, \hat{\theta})_L = \argmax_{\phi, \theta} \frac{1}{M^2} \sum_{i=1}^{M^2} \hat{\mathcal{I}}_{\phi, \theta}(C^{(i)}_{\phi}(X); E_{\phi}(X)),
\end{equation}
where $C^{(i)}_{\phi}(X)$ is a feature map location from encoder (with a limited receptive field corresponding to an input patch with size $M$) at some intermediate layer of the network.

Due to stronger performance of AlexNet architecture (Section~\eqref{Supervised_baselines}) in our experiments (see Section~\eqref{Results}) we used it as an encoder for DIM method. Last linear layer of AlexNet we changed with a layer for $64$-dimensional output representation.

To estimate mutual information using eq.~\eqref{eq:diim-l} we used the encode-and-dot-product architecture (Fig.~$6$ from \cite{hjelm2018learning}). First, patches $C^{(i)}_{\phi}(X)$ taken from third convolutional layer of AlexNet were mapped using convolutional encoder-and-dot architecture (Tab.~$9$ from \cite{hjelm2018learning}) with $512$ units and their representation $Z = E_{\phi}(X)$ --- linear encoder-and-dot architecture (Tab.~$8$ from \cite{hjelm2018learning}). Then flattened encoded mappings of patches and representations were combined using the dot product to create real and fake samples efficiently. The real sample is a dot product of a ``local" patch and its ``global" representation mappings, while fake --- between mapping of some ``local" patch with global representation coming from an unrelated input. Eventually we estimated JSD based loss eq.~\eqref{eq:diim_jsd} and NCE --- eq.~\eqref{eq:diim_nce} using these samples. Since NCE needs to have more negative samples to be competitive with JSD~\cite{hjelm2018learning}, all possible combinations between the patch and representation mappings were used a similar way to create negative samples.

To evaluate the performance of the learned representation by DIM, we trained three additional neural networks using as input features output from last convolutional layer with size $128 \times 2 \times 2 \times 2$, the first fully connected layer with $1024$ units, and final fully connected layer with $64$-dimensional representation, which we call as Conv, FC, and $Z$.
The classifiers are composed of one fully-connected layer with $200$ hidden units, dropout~\cite{srivastava2014dropout} with $p=0.1$, batch normalization~\cite{ioffe2015batch} and a ReLU~\cite{nair2010rectified}  activation.

\begin{table}[ht]
    \centering
    \small
    \caption{AlexNet and ResNet architectures}
    \begin{tabular}{|l|c|c|c|}
    \hline
    {\bf AlexNet}\\
    \hline
    3D Conv $(1,64,5,2,0)$ - BN 3D - ReLU - MP 3D $(3,3)$ \\
    3D Conv $(64,128,3,1,0)$ - BN 3D - ReLU - MP 3D $(3,3)$ \\
    3D Conv $(128,192,3,1,1)$ - BN 3D - ReLU \\
    3D Conv $(192,192,3,1,1)$ - BN 3D - ReLU \\
    3D Conv $(192,128,3,1,1)$ - BN 3D - ReLU - MP 3D $(3,3)$ \\
    Linear $(1024,1024)$ - BN 1D - ReLU\\
    Linear $(1024,4)$ - SoftMax - ArgMax \\
    \hline
    {\bf ResNet}\\
    \hline
    3D Conv $(1, 64, 3, 2, 0)$ - BN 3D - ReLU - MP 3D $(3,3)$\\
    Residual Layer 1 \\
    \tab BB0 - 2 x (3D Conv $(64, 64, 3, 1, 1)$ - BN 3D - ReLU)\\
    \tab BB1 - 2 x (3D Conv $(64, 64, 3, 1, 1)$ - BN 3D - ReLU)\\
    Residual Layer 2\\
    \tab BB0 - 3D Conv $(64, 128, 3, 2, 1)$ - BN 3D - ReLU\\
    \tab BB0 - 3D Conv $(128, 128, 3, 1, 1)$ - BN 3D - ReLU\\
    \tab BB0 downsample - 3D Conv $(64, 128, 3, 2, 1)$ - BN 3D\\
    \tab BB1 - 2 x (3D Conv $(128, 128, 3, 2, 1)$ - BN 3D - ReLU)\\
    Residual Layer 3\\
    \tab BB0 - 3D Conv $(128, 256, 3, 2, 1)$ - BN 3D - ReLU\\
    \tab BB0 - 3D Conv $(256, 256, 3, 1, 1)$ - BN 3D - ReLU\\
    \tab BB0 downsample - 3D Conv $(128, 256, 3, 2, 1)$ - BN 3D\\
    \tab BB1 - 2 x (3D Conv $(256, 256, 3, 2, 1)$ - BN 3D - ReLU)\\
    MaxPool 3D $(3, 3)$\\
    Linear $(2048,1024)$ - BN 1D - ReLU\\
    Linear $(1024,4)$ - SoftMax - ArgMax \\
    \hline
    \end{tabular}
\label{tb:cnn_arc}
\end{table}

\subsection{Supervised baselines} \label{Supervised_baselines}
As baselines we have considered supervised methods --- two convolutional networks, one based on a simplified AlexNet~\cite{NIPS2012_4824} architecture and the other a ResNet~\cite{he2016deep} architecture. Both networks use convolutions and max pooling with volumetric kernels, batch normalization, ReLU and two fully connected layers in the end (see Tab.~\eqref{tb:cnn_arc} for details). The notations in Tab.~\eqref{tb:cnn_arc} denotes: \emph{BN} for batch normalization, \emph{BB} --- a basic block, \emph{MP $(k, s)$} ---  max pooling with kernel size $k$ and stride $s$, for convolutions \emph{$(i,o,k,s,p)$} --- a number of input and output channels, a kernel size, a stride and a padding respectively). Cross-entropy loss used as a training objective.

\subsection{Regularization}

For small datasets, it is common to penalize the number of the model parameters by driving most of them to zero using $L_1$ regularization. Formally, this penalty is defined as:
\begin{equation}
        L_1 (\omega) =  \lambda ||\omega||_1 = \lambda \sum_i |\omega_i|,
\end{equation}
where $\omega$ is parameter vector of the model and $\lambda$  --- coefficient. $L_1$ regularization imposes a sparse solution. This penalty is added to JSD, NCE and cross-entropy losses in different setting. For our experiments we used $\lambda = 1$.

\section{Experiments}

\subsection{Datasets and preprocessing}
For the downstream classification task, the data was obtained from the ADNI database adni.loni.usc.edu (for up-to-date information, see www.adni-info.org).
We use T1w MRI images of $830$ subjects with four different groups: patients with stable, and progressive MCI, Alzheimer's disease and healthy controls. 

Structural MRI (sMRI) data was pre-processed to grey matter volume (modulated) maps using SPM12 toolbox. 
To segment grey matter, the MRI images were spatially normalized and smoothed by 6 mm full width at half maximum (FWHM) 3D Gaussian kernel. 
After quality control, two subjects from ADNI dataset were excluded.
The final dataset consisted of $828$ subjects with a volume size of $[121, 145, 121]$.

\subsection{Experimental setup}

\subsubsection{Data}
The dataset was divided in approximately $93\%$ and $7\%$ subjects for cross-validation and hold-out test sets using a stratified split. 
Then, $93\%$ subjects were split into five stratified folds.

For AlexNet and ResNet architectures, we used simple data augmentation of the training dataset to reduce overfitting to the small number of annotated samples available. 
Our augmentation consisted of zero padding and random cropping to size $128$ along all dimensions along with randomly flipping the input with probability $0.5$ for each axis. The whole brain was included in the crop. 

For DIM, we didn't use data augmentation, but we used zero padding to make sure that input size is equal to $128$ along all dimensions.

\subsubsection{Training}
The models were trained using the AMSGrad~\cite{reddi2019convergence} optimizer with learning rate $0.001$ for CNN models and $0.0001$ for DIM using a batch size of $8$ but dropping the incomplete last batch.
The training of the supervised architectures was performed for $500$ epochs, 
DIM --- for $1000$ epochs as pre-training and for $1000$ epochs for training the classifiers on top of frozen features from the encoder.  

\subsubsection{Evaluation}
Since the dataset is not completely balanced, the evaluation was performed using balanced accuracy~\cite{brodersen2010balanced}, defined as the average of recall of each class (implementation in scikit-learn~\cite{scikit-learn}).

\subsubsection{Implementation and hardware}
The implementation was written using Deep Learning frameworks PyTorch~\cite{paszke2017automatic} and Cortex~\cite{cortex}. The DIM code is based on openly available DIM implementation~\cite{dim_code}. The experiments were performed on NVIDIA GeForce Titan X Pascal and 1080 Ti and 8 CPU threads.

\begin{figure}[ht]
\centerline{\includegraphics[width=\linewidth,keepaspectratio]{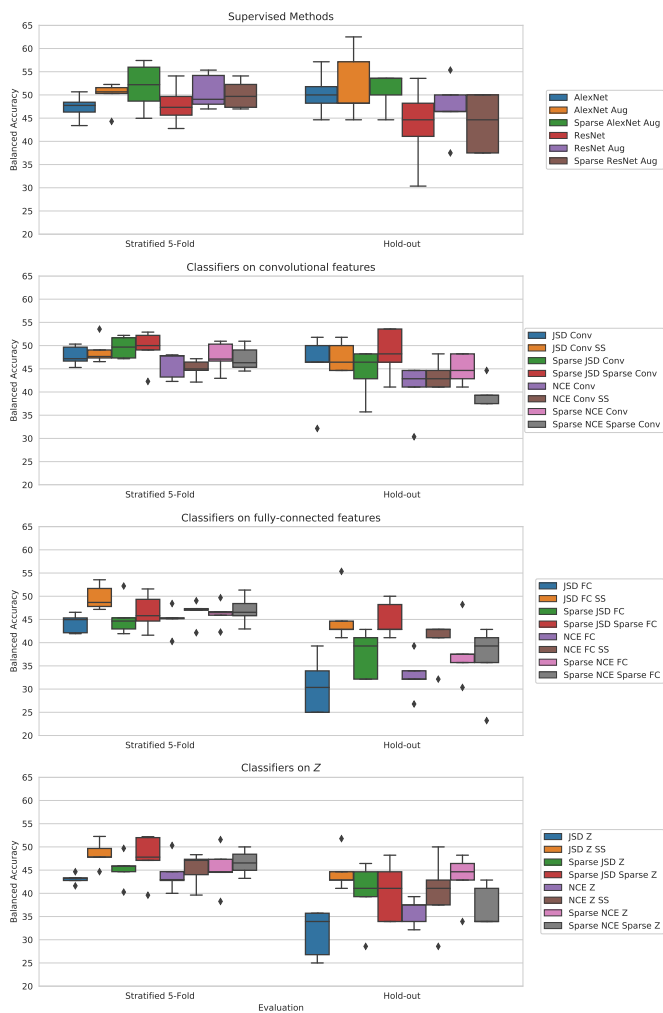}}
\caption{Performance of the models}
\label{boxplot_perf}
\end{figure}

\begin{table}[ht]
    \small
    \label{tb:reults_bal_acc}
    \centering
    \caption{Performance}
    \begin{adjustbox}{width=\columnwidth,center}
    \begin{tabular}{|l|c|c|c|c|c|}
        \hline
        \multirow{2}{*}{\textbf{Model}} & \multirow{2}{*}{\textbf{\begin{tabular}[c]{@{}l@{}}Balanced Accuracy\\ Stratified 5-Fold\end{tabular}}} & \multirow{2}{*}{\textbf{\begin{tabular}[c]{@{}l@{}}Balanced Accuracy\\ Hold-out\end{tabular}}} & \textbf{Mean gap} &  \multicolumn{2}{l|}{\textbf{Wilcoxon test}} \\ \cline{5-6} 
        & & & & Stat & $p$-value \\ \hline
        AlexNet & $47.31 \pm 2.69$ & $50.36 \pm 4.62$ & $3.05$& 6.5 & \boldmath$0.290$\\
        AlexNet Aug & $49.82 \pm 3.18$ & \boldmath$52.14 \pm 7.41$ & $2.32$& 7.0 & \boldmath$0.554$\\
        \textbf{Sparse AlexNet Aug} & \boldmath$51.85 \pm 5.14$ & $51.07 \pm 3.91$ & \boldmath$0.78$& N/A & N/A\\
        ResNet & $47.9 \pm 4.28$ & $43.57 \pm 8.71$ & $4.33$& 10.0 & 0.034\\
        ResNet Aug & $50.72 \pm 3.8$ & $47.14 \pm 6.51$ & $3.58$& 14.0 & 0.039\\
        Sparse ResNet Aug & $50.07 \pm 3.09$ & $43.93 \pm 6.26$ & $6.14$& 10.0 & 0.033\\ 
        \hline
        JSD Conv & $47.82 \pm 2.11$ & $45.36 \pm 7.74$ & $2.46$& 13.5 & \boldmath$0.052$\\
        JSD Conv SS & $48.83 \pm 2.79$ & $47.5 \pm 3.24$ & $1.33$& 13.5 & \boldmath$0.052$\\
        Sparse JSD Conv & \boldmath$49.61 \pm 2.35$ & $44.29 \pm 5.27$ & $5.32$& 15.0 & 0.022\\
        \textbf{Sparse JSD Sparse Conv} & $49.29 \pm 4.22$ & \boldmath$48.57 \pm 5.27$ & \boldmath$0.72$& 6.0 & \boldmath$0.054$\\
        NCE Conv & $45.82 \pm 2.82$ & $40.71 \pm 5.98$ & $5.11$& 15.0 & 0.022\\
        NCE Conv SS & $45.08 \pm 1.94$ & $43.57 \pm 2.99$ & $1.51$& 10.0 & 0.034\\
        Sparse NCE Conv & $47.59 \pm 3.21$ & $45.0 \pm 3.19$ & $2.59$& 15.0 & 0.022\\
        Sparse NCE Sparse Conv & $47.23 \pm 2.69$ & $39.64 \pm 2.93$ & $7.59$& 15.0 & 0.022\\
        \hline
        JSD FC & $44.18 \pm 2.05$ & $30.71 \pm 6.11$ & $13.47$& 15.0 & 0.022\\
        \textbf{JSD FC SS} & \boldmath$49.77 \pm 2.73$ & \boldmath$45.36 \pm 5.73$ & $4.41$& 14.0 & 0.040\\
        Sparse JSD FC & $45.42 \pm 4.03$ & $37.5 \pm 5.05$ & $7.92$& 15.0 & 0.022\\
        Sparse JSD Sparse FC & $46.6 \pm 3.92$ & $45.0 \pm 3.87$ & \boldmath$1.6$& 15.0 & 0.022\\
        NCE FC & $44.89 \pm 2.93$ & $32.86 \pm 4.48$ & $12.03$& 15.0 & 0.022\\
        NCE FC SS & $46.53 \pm 2.59$ & $40.36 \pm 4.66$ & $6.17$& 15.0 & 0.020\\
        Sparse NCE FC & $46.2 \pm 2.64$ & $37.86 \pm 6.49$ & $8.34$& 15.0 & 0.022\\
        Sparse NCE Sparse FC & $47.01 \pm 3.12$ & $36.43 \pm 7.84$ & $10.58$& 15.0 & 0.022\\
        \hline
        JSD Z & $43.12 \pm 1.1$ & $31.43 \pm 5.14$ & $11.69$& 15.0 & 0.021\\
        \textbf{JSD Z SS} & \boldmath $48.44 \pm 2.79$ & \boldmath$44.64 \pm 4.19$ & $3.8$& 15.0 & 0.022\\
        Sparse JSD Sparse Z & $47.74 \pm 5.12$ & $40.36 \pm 6.39$ & $7.38$& 15.0 & 0.022\\
        Sparse JSD Z & $45.27 \pm 3.38$ & $40.0 \pm 6.98$ & $5.27$& 15.0 & 0.022\\
        NCE Z & $44.14 \pm 3.84$ & $36.07 \pm 2.93$ & $8.07$& 15.0 & 0.022\\
        NCE Z SS & $45.28 \pm 3.55$ & $40.0 \pm 7.84$ & $5.28$& 14.0 & 0.040\\
        Sparse NCE Sparse Z & $46.63 \pm 2.69$ & $37.14 \pm 4.45$ & $9.49$& 15.0 & 0.022\\
        Sparse NCE Z & $45.27 \pm 4.85$ & $43.21 \pm 5.56$ & \boldmath$2.06$& 15.0 & 0.021\\
        \hline
        \end{tabular}
    \end{adjustbox}

\end{table}

\section{Results} \label{Results}

The final trained models used further to evaluate the performance were selected based on the best-balanced accuracy but from a checkpoint where the validation score was lower than the training score. We gave the model a burn-in period before applying this rule to deal with initial stochasticity. The models notations are as follows: \emph{Aug} denotes augmentation of the training dataset, the first \emph{sparse} --- a model trained with $L_1$ regularization, the second --- a classifier on top of the frozen features from encoders trained using $L_1$ regularization, \emph{SS} --- stands for training an unsupervised model with an additional supervised loss from $Z$-classifier.

Table~\ref{tb:reults_bal_acc} reports the balanced accuracy rates including mean, standard deviation values, and the gap between mean values on cross-validation and hold-out.  The bold text distinguishes the best scores and the name of the models. The last column shows $p$-value and statistic for the one-sided Wilcoxon test. The bold $p$-values indicate acceptance of the null hypothesis. The test was performed to compare each method with the best model (\emph{Sparse  AlexNet   Aug}) based on the five values of balanced accuracy on hold-out. An alternative hypothesis is that the model Sparse AlexNet Aug is better. Fig.~\ref{boxplot_perf} highlights the distributions of the performance.

With all modifications, \emph{ResNet} shows a lower performance on hold-out (at most $47.14 \pm 6.51$) than \emph{AlexNet}. It is reasonable since the capacity of the ResNet architecture is larger and the dataset is small. For $\alpha=0.05$ Wilcoxon test also rejects H0 supporting the worse performance of ResNet. Performance of \emph{JSD Conv}, \emph{JSD Conv SS}, \emph{Sparse  JSD  Sparse  Conv}, \emph{AlexNet}, \emph{AlexNet Aug} is statistically indistinguishable from that of \emph{Sparse  AlexNet  Aug}. Follows that unsupervised DIM has comparable performance to supervised methods.

Among DIM variants, JSD has higher scores than NCE. Lower scores of NCE can be explained by its requirement of a large number of negative samples during training to be competitive with JSD. Our dataset is not large enough to support the needed level of negative sampling. 

The best score with convolutional features---$48.57 \pm 5.27$---was obtained by an encoder and classifier trained with $L_1$ regularization which is the \emph{Sparse JSD Sparse Conv} model. For features from the fully-connected layer --- \emph{JSD FC SS} model with $45.36 \pm 5.73$  using semi-supervised loss was the best. However, \emph{Sparse JSD Sparse FC} has similar results $45.0 \pm 3.87$ and a smaller mean gap $1.6$ but it has a lower mean cross-validation score by $3.17\%$ . For the smallest $64$-dimensional representation, semi-supervised model \emph{JSD Z SS} gives the best performance $44.64 \pm 4.19$, but similar result $43.21 \pm 5.56$ were obtained by \emph{Sparse NCE Z} model. Semi-supervised loss and $L_1$ regularization improved models' generalization by reducing the gap between cross-validation and hold-out scores. The observed degradation in performance between \emph{Conv}, \emph{FC}, and $Z$ can be explained by the reduced capacity of the features. $L_1$ regularization and dropout could also be adjusted. However, a more compact input representation can be of independent use, for example, for dimensionality reduction.

In previous studies, the best reported accuracy for the ResNet architecture in a 4-class sMRI classification task was $54\%$~\cite{abrol2018deep}, while stacked autoencoders (SAE)~\cite{6963480} reached for sMRI only $46.30 \pm 4.24$ and for sMRI+PET $53.79 \pm 4.76$, and DW-S$^2$MTL~\cite{suk2016deep} --- for sMRI $47.83$ or for sMRI+PET+CSF $53.72$. Our values can't be completely comparable since the evaluation is different. Reproduced ResNet can be used as a proxy to estimate performance relative to this prior work. Note, however, it is not one of the best-performing methods in our study.

\section{Conclusions}

This work proposes an unsupervised method DIM for learning representations from structural neuroimaging data. The evaluation of the prediction of progression to Alzheimer's disease demonstrates results comparable to supervised methods. In the future, we will scale up our experiments with increased sample size and address the cases of other diseases. Our future efforts will also be focused on the multi-modal fusion of brain imaging data~\cite{calhoun2016multimodal} to increase the predictive strength of the model.

\section*{Acknowledgement}

This study is supported by NIH grants R01EB020407, R01EB006841, P20GM103472, P30GM122734.

Data collection and sharing for this project was funded by the Alzheimer's Disease Neuroimaging Initiative (ADNI) (National Institutes of Health Grant U01 AG024904) and DOD ADNI (Department of Defense award number W81XWH-12-2-0012). ADNI is funded by the National Institute on Aging, the National Institute of Biomedical Imaging and Bioengineering, and through generous contributions from the following: AbbVie, Alzheimer's Association; Alzheimer's Drug Discovery Foundation; Araclon Biotech; BioClinica, Inc.; Biogen; Bristol-Myers Squibb Company; CereSpir, Inc.; Cogstate; Eisai Inc.; Elan Pharmaceuticals, Inc.; Eli Lilly and Company; EuroImmun; F. Hoffmann-La Roche Ltd and its affiliated company Genentech, Inc.; Fujirebio; GE Healthcare; IXICO Ltd.;Janssen Alzheimer Immunotherapy Research \& Development, LLC.; Johnson \& Johnson Pharmaceutical Research \& Development LLC.; Lumosity; Lundbeck; Merck \& Co., Inc.;Meso Scale Diagnostics, LLC.; NeuroRx Research; Neurotrack Technologies; Novartis Pharmaceuticals Corporation; Pfizer Inc.; Piramal Imaging; Servier; Takeda Pharmaceutical Company; and Transition Therapeutics. The Canadian Institutes of Health Research is providing funds to support ADNI clinical sites in Canada. Private sector contributions are facilitated by the Foundation for the National Institutes of Health (www.fnih.org). The grantee organization is the Northern California Institute for Research and Education, and the study is coordinated by the Alzheimer's Therapeutic Research Institute at the University of Southern California. ADNI data are disseminated by the Laboratory for Neuro Imaging at the University of Southern California.

\bibliographystyle{IEEEtran}
\bibliography{references}

\end{document}